\documentclass[preprint,12pt,authoryear]{elsarticle}
\usepackage{natbib}  
\usepackage{algorithmic}
\usepackage{diagbox}  
\usepackage{pifont}
\usepackage{amsmath,graphicx,hyperref,amsfonts}
\newcommand{\cxmark}{\ding{55}}
\usepackage{threeparttable} 
\usepackage{xcolor}
\usepackage{amssymb}
\usepackage{mathrsfs}
\usepackage{multirow}
\usepackage[T1]{fontenc}

\def\BibTeX{{\rm B\kern-.05em{\sc i\kern-.025em b}\kern-.08em
    T\kern-.1667em\lower.7ex\hbox{E}\kern-.125emX}}

\journal{ }

\begin{document}

\begin{frontmatter}
\title{$\text{SSA}^3\text{D}$: Text-Conditioned Assisted Self-Supervised Framework for Automatic Dental Abutment Design}

\author[label1,label2,label3]{Mianjie Zheng}
\ead{2400101029@mails.szu.edu.cn}
\author[label1,label2,label3]{Xinquan Yang\corref{mycorrespondingauthor}}
\ead{scxxy2@nottingham.edu.cn}
\author[label3,label4]{Along He}
\ead{healong2020@163.com}
\author[label5]{Xuguang Li}
\ead{lixuguang@szu.edu.cn}
\author[label7]{Feilie Zhong}
\ead{phillip@ktjdental.com}
\author[label1,label2,label3]{Xuefen Liu}
\ead{2400101034@mails.szu.edu.cn}
\author[label1,label2,label3]{Kun Tang}
\ead{2500101065@mails.szu.edu.cn}
\author[label6]{Zhicheng Zhang}
\ead{zhangzhicheng13@mails.ucas.edu.cn}
\author[label1,label2,label3,label4]{Linlin~Shen\corref{mycorrespondingauthor}}
\ead{llshen@szu.edu.cn}
\cortext[mycorrespondingauthor]{Corresponding author}

\address[label1]{College of Computer Science and Software Engineering, Shenzhen University, Shenzhen, China}
\address[label2]{AI Research Center for Medical Image Analysis and Diagnosis, Shenzhen University, Shenzhen, China}
\address[label3]{National Engineering Laboratory for Big Data System Computing Technology, Shenzhen University, China}
\address[label4]{School of Artificial Intelligence, Shenzhen University, Shenzhen, China}
\address[label5]{Department of Stomatology, Shenzhen University General Hospital, Shenzhen, China}
\address[label6]{Shenzhen Institute of Advanced Technology, Chinese Academy of Science, Shenzhen, China}
\address[label7]{Kangtaijian (KTJ) Medical Technology Co., Ltd., Shenzhen, China}


%
\begin{abstract}
Abutment design is a critical step in dental implant restoration. However, manual design involves tedious measurement and fitting, and research on automating this process with AI is limited, due to the unavailability of large annotated datasets. Although self-supervised learning (SSL) can alleviate data scarcity, its need for pre-training and fine-tuning results in high computational costs and long training times.
In this paper, we propose a Self-supervised assisted automatic abutment design framework (SS$A^3$D), which employs a dual-branch architecture with a reconstruction branch and a regression branch. The reconstruction branch learns to restore masked intraoral scan data and transfers the learned structural information to the regression branch. The regression branch then predicts the abutment parameters under supervised learning, which eliminates the separate pre-training and fine-tuning process.
We also design a Text-Conditioned Prompt (TCP) module to incorporate clinical information (such as implant location, system, and series) into SS$A^3$D. This guides the network to focus on relevant regions and constrains the parameter predictions. 
Extensive experiments on a collected dataset show that SS$A^3$D saves half of the training time and achieves higher accuracy than traditional SSL methods. It also achieves state-of-the-art performance compared to other methods, significantly improving the accuracy and efficiency of automated abutment design.
\end{abstract}
\begin{keyword}
Dental Abutment, Text localization, Dental implant, Regression, Self-supervised learning
\end{keyword}
\end{frontmatter}
\section{Introduction}\label{introduction}
\begin{figure*}\centering
\includegraphics[scale=0.35]{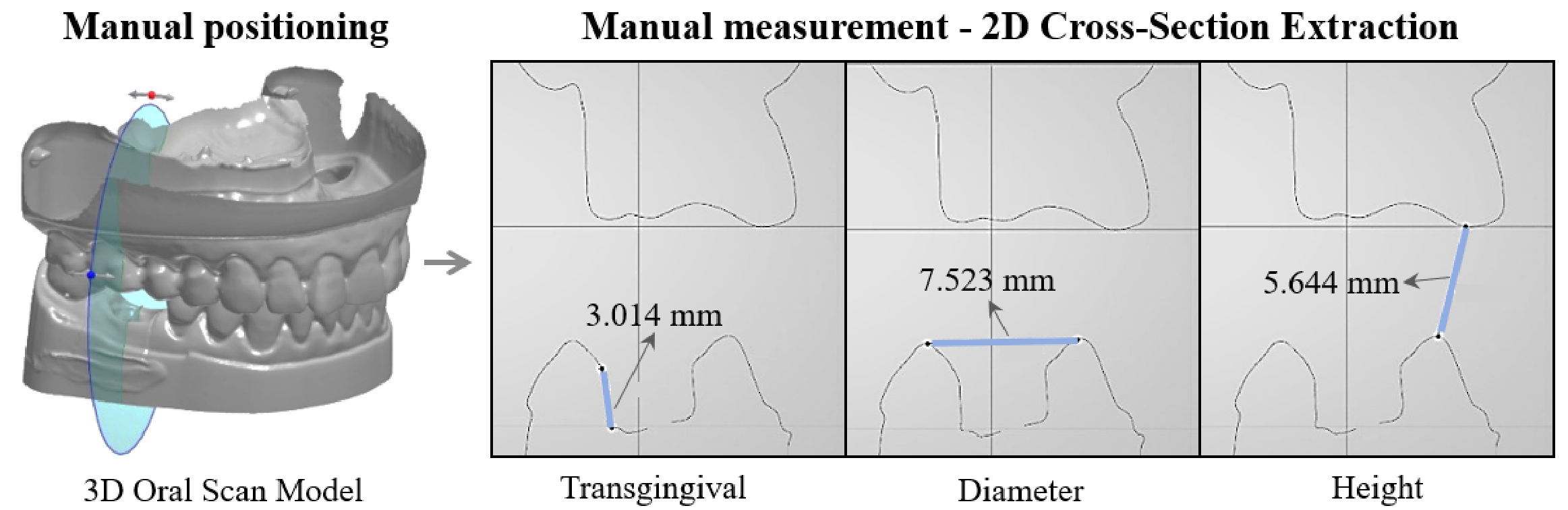}
\caption{Manual abutment design process based on CAD-CAM technology. 
}
\label{process_framework}
\end{figure*}

Artificial dental implantation is one of the most appropriate treatment methods for tooth loss. The dental abutment is an important component of the dental implant, serving as the intermediate structure connecting the implant to the superstructure. Its core functions include providing mechanical support to the crown, enhancing the retention of the restoration, and maintaining long-term occlusal stability. 
Traditionally, the restoration space for implant abutments has been measured either manually or through computer-assisted design and computer-assisted manufacturing (CAD-CAM) techniques~\cite{zhang2017method}. 
The manual method requires printing the 3D model of the patient's mouth and manually measuring the restoration space of the abutment, including the thickness of the oral gingiva (transgingival), the diameter of the implant position (diameter), and the gingival-mandibular distance (height).
CAD-CAM-based methods employ digital scanning to measure the restoration space~\cite{tartea2023comparative, benakatti2021dental}, facilitating the design of abutments with enhanced precision. 
Although CAD-CAM application eliminates the need for physical printing and repeated fitting, the measurement process still necessitates manual involvement (Fig.~\ref{process_framework} illustrates the process for determining these three parameters), making the overall workflow time-consuming and labor-intensive~\cite{shah2023literature}. 
Furthermore, as manual measurements rely heavily on subjective judgment, they are prone to numerical deviations. 
Such inaccuracies can jeopardize long-term implant stability, as poorly fitted abutments may lead to biological complications such as peri-implantitis \cite{choi2023influence}.

These limitations of traditional and CAD-CAM-assisted manual design highlight a pressing need for a paradigm shift that can minimize subjective intervention and enhance precision. Coincidentally, the field of artificial intelligence (AI) has recently demonstrated remarkable capabilities in automating complex design and measurement tasks across various domains.
In dentistry, AI has already achieved breakthroughs in tasks requiring high spatial and morphological precision, such as tooth segmentation~\cite{cui2021tsegnet,jang2021fully,qiu2022darch,xu20183d}, crown restoration~\cite {tian2021efficient,shen2023transdfnet,yuan2020personalized}, and dental implant planning~\cite{yang2024two,yang2024implantformer,yang2024simplify}).
The success of AI in these related areas suggests its strong potential to address the very challenges inherent in abutment design. 
By leveraging AI to interpret digital scans and automatically generate design parameters, the process could transition from a manual, skill-dependent task to an automated, objective, and data-driven one. 

Generally, training models for automated abutment design necessitates large-scale labeled data. 
However, due to the difficulty in obtaining abutment parameters, it is impractical to collect a large amount of labeled data for network training.
The scarcity of labeled data has impeded the progress of artificial intelligence. 
Lately, self-supervised learning (SSL) has achieved great success in deep learning.
By extracting meaningful and general-purpose representations from unlabeled data through pretext tasks, the supervision signals are automatically derived from the intrinsic structure or correlations within the data. 
These representations can then be transferred to downstream tasks, effectively mitigating the limitations imposed by scarce labeled data and enhancing the performance of downstream tasks with small amounts of data \cite{krishnan2022self}.  
Although SSL methods have been successfully applied to 3D dental tasks \cite{liu2022hierarchical,krenmayr2025evaluating,almalki2024self,ma2025multi}, it typically involves prolonged pretraining followed by fine-tuning, making the overall process cumbersome and resource-intensive.
To overcome these limitations, recent studies have proposed self-supervised auxiliary tasks (SSAT). SSAT adopts a multi-branch network framework that integrates SSL during training to learn representations and share them with downstream tasks, while only the downstream branch is activated during inference. This design eliminates redundant fine-tuning steps, enables direct end-to-end optimization, reduces computational cost and training time, and achieves superior performance compared with conventional SSL methods~\cite{das2024limited, liu2021efficient}.
The success of the SSAT paradigm inspired us to develop an automatic abutment design framework, which can learn more robust features on limited data. 

\begin{figure*}\centering
\includegraphics[scale=0.27]{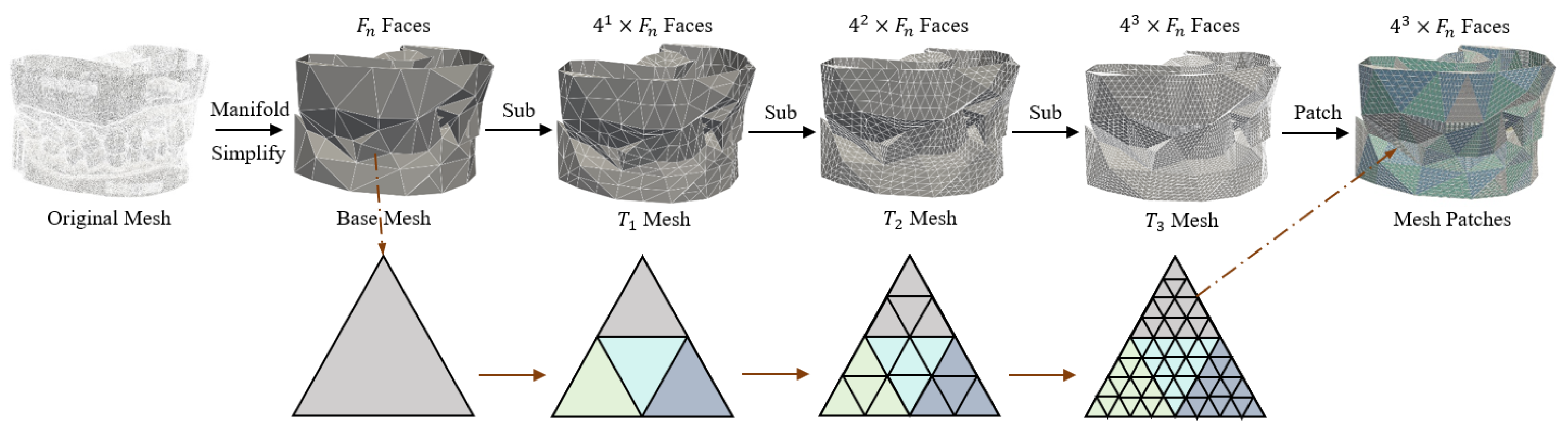}
\caption{Visualization of the remesh processing. First, the original oral scan mesh data are converted into a base mesh through manifold and simplification operations. The base mesh is then refined three times, with each refinement step subdividing every face into four smaller ones. Finally, the refined mesh is used as the network input.}
\label{remesh_ope}
\end{figure*}

In this paper, we propose a self-supervised assisted automatic abutment design framework (SS$A^3$D), which simplifies the abutment design process and performs well with limited labeled data.
The architecture of the proposed SS$A^3$D is shown in Fig.~\ref{Overall_Framework}. 
It integrates a reconstruction branch and a regression branch via a shared mesh encoder.
The reconstruction branch extracts mesh features from masked oral scan data and uses a decoder to restore the texture, enabling the encoder to learn oral structure information. 
The regression branch is responsible for the prediction of the abutment parameter from complete intraoral scan data. 
Clinically, the information on the implant system and implant position is important for the determination of abutment. 
Dentists will determine this information before designing the abutment.
To integrate this prior knowledge into SS$A^3$D, we design a text-conditioned prompt (TCP) module.
First, we incorporate both implantation system/series information and implant location into textual descriptions, and then the CLIP text encoder is used to encode textual descriptions and generate the corresponding text embeddings.
The TCP module then fused the text embeddings with the output feature of the mesh encoder via cross-attention, guiding SS$A^3$D to localize implant regions and determine abutment parameters.
We validate the proposed framework on a collected abutment design dataset. 
The main contributions of this paper are as follows:
\begin{itemize}
\item We propose SS$A^3$D, a assisted self-supervised framework for automatic abutment design, exhibiting superior performance compared to the SSL-based method on limited training data. 
\item A text-conditioned prompt (TCP) module is proposed to introduce clinical prior knowledge into SS$A^3$D to constrain parameter regression and guide implant localization.
\item Extensive experiments are conducted on a collected abutment design dataset, demonstrating that the proposed SS$A^3$D has higher efficiency and achieves state-of-the-art performance.
\end{itemize}

\section{Related Work}
\subsection{Traditional Abutment Design}
Measuring the restoration space is a critical step in abutment design, as it directly influences the fit between the abutment and the patient’s oral structure \cite{abichandani2013abutment}. 
In clinical practice, the manual abutment design process is both cumbersome and time-consuming \cite{shah2023literature}. 
Dentists must scan the patient’s intraoral structures, create gingival models, determine the implantation system information, measure the restoration space, preselect and install the abutment, verify compatibility, and finally confirm the most suitable abutment. 
To streamline the process, CAD technology has been introduced for parameter measurement and abutment design on digital models, eliminating the need for physical printing and repeated manual verification \cite{priest2005virtual, ramadan2025fully}.
Moreover, CAD-CAM technology enables the design of customized abutments, providing a better fit and higher personalization while avoiding dimensional inaccuracies inherent in traditional waxing, investing, and casting processes \cite{zhang2017method}. The application of CAD-CAM reduces material waste and improves design efficiency \cite{kang2020abutment}.
Nevertheless, these approaches also present limitations: both manual and CAD-CAM-based methods rely on positioning and manual measurement, which can require resolution by experienced experts. 
Selecting different positions along the gingival contour during measurement can lead to inconsistent outcomes, potentially resulting in inappropriate abutments and biological complications during long-term implant placement \cite{prpic2025influence}.

\subsection{Self-Supervised Learning and Self-Supervised Auxiliary Tasks}
Self-supervised learning (SSL) alleviates the problem of limited annotated data by learning meaningful feature representations through pretext tasks. Typical approaches include contrastive learning (CL) and Masked Image Modeling (MIM). CL-based methods minimize the distance between augmented views of the same sample while maximizing the distance between samples of different categories \cite{chen2020simple, chen2021exploring, he2020momentum}.  
Zhang et al. proposed the CMAE-3D framework and designed Hierarchical Relation Contrastive Learning (HRCL) to mine semantic similarity information from the voxel level and frame level, effectively alleviating the negative sample mismatch problem in contrastive learning \cite{zhang2025cmae}.
On the other hand, MIM-based methods learn representations by randomly masking portions of the input and reconstructing the missing content \cite{bao2021beit, xu2024self}.  
Wei et al. introduced Latent MIM, which replaces pixel-level reconstruction with latent-space reconstruction to overcome key limitations of traditional MIM, including representation collapse, reconstruction target selection \cite{wei2024towards}.
Overall, an increasing number of SSL methods have demonstrated strong capability in representation learning. 
\par
Although SSL methods exhibit good performance in many tasks, it typically involves prolonged pretraining followed by fine-tuning, making the overall process cumbersome and resource-intensive.
Self-supervised auxiliary task (SSAT) jointly optimizes SSL and downstream tasks in an end-to-end manner, simplifying training and achieving better performance than SSL. 
Liu et al.~\cite{liu2021efficient} proposed a dense relative localization auxiliary self-supervised task, which samples pairs of token embeddings, predicts their normalized 2D relative distances through an MLP, and combines this with cross-entropy loss for supervision. 
Das et al.~\cite{das2024limited} developed an SSAT framework that jointly optimizes the primary classification task with a MAE-based reconstruction task, rather than following the conventional two-stage SSL pre-training and fine-tuning pipeline. 
Recently, SSAT has also been applied in dentistry. Cai et al.~\cite{cai2024ssad} proposed SSAD, 
which integrates a simMIM-based image reconstruction branch with a disease detection branch. SSAD outperforms mainstream SSL-based methods while reducing training time by up to 8.7 hours. 

\subsection{Self-Supervised Learning in Oral Scan Data}
Oral scan data is mainly represented in two forms: point cloud and mesh. 
SSL-based methods have been increasingly applied to the oral point cloud and mesh data.
Liu et al. proposed STSNet~\cite{liu2022hierarchical}, a hierarchical SSL framework for 3D tooth segmentation, which leverages unlabeled data for unsupervised pre-training through point-level, region-level, and cross-level contrastive losses, achieving superior performance compared with fully supervised baselines using only limited labeled data. 
Krenmayr et al.~\cite{krenmayr2025evaluating} evaluated four masked SSL frameworks—Point-BERT \cite{yu2022point} (integrating the BERT paradigm with a discrete variational autoencoder tokenizer and auxiliary contrastive learning), Point-MAE~\cite{pang2022masked} (employing a lightweight PointNet encoder with a shared learnable decoder token), Point-GPT~\cite{chen2023pointgpt} (using Morton-order geometric encoding with dual masking), and Point-M2AE~\cite{zhang2022point} (featuring a multi-scale pyramid encoder-decoder with progressive masking) — for 3D dental model segmentation, and verified the effectiveness of SSL in this domain.
In mesh data, Almalki et al. proposed DentalMAE~\cite{almalki2024self}, which uses a ViT encoder and a lightweight decoder pre-trained on a 3D teeth dataset and fine-tuned for tooth segmentation, outperforming fully supervised methods and demonstrating strong generalization on limited data.
Ma et al. introduced MECSegNet~\cite{ma2025multi}, a dual-stream SSL segmentation network based on multi-encoding contrastive learning, where encoders are pre-trained with local, global, and cross-stream contrastive losses. MECSegNet not only outperforms existing SSL methods but also reduces computational cost by 79.81\%. 
Collectively, these methods demonstrate the potential of SSL in oral scan data.

\begin{figure*}\centering
\includegraphics[scale=0.23]{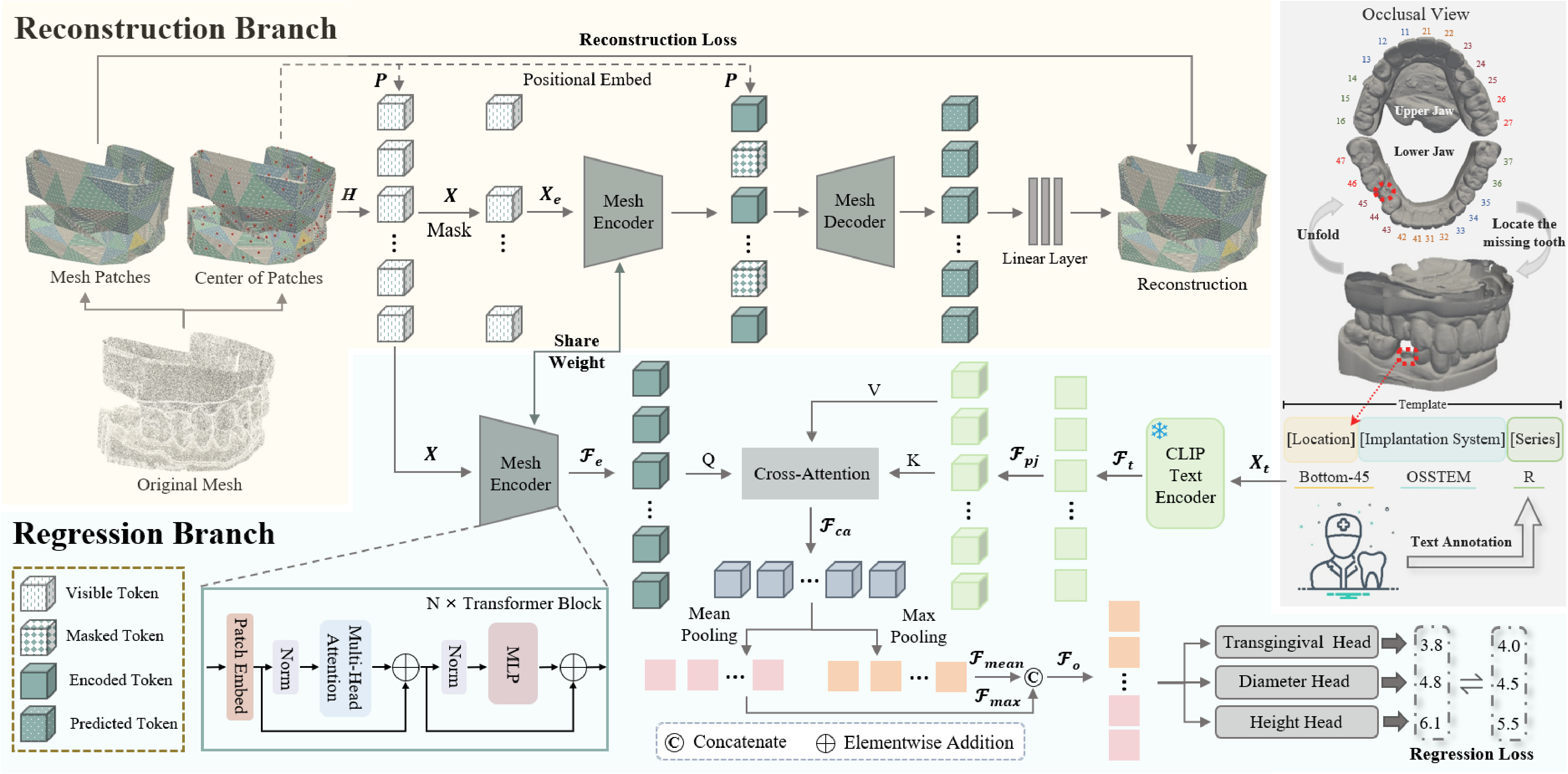}
\caption{The overall framework of the proposed SS$A^3$D method.}
\label{Overall_Framework}
\end{figure*}

\section{Method} \label{method}
The architecture of the proposed SS$A^3$D is given in Fig.~\ref{Overall_Framework}, which integrates a reconstruction branch and a regression branch via a shared mesh autoencoder.
The reconstruction branch extracts mesh features from masked oral scan data and uses a decoder to restore the texture, enabling the autoencoder to learn oral structure information. 
The regression branch is responsible for the prediction of the abutment parameter from the input oral scan data. 
Next, we will introduce them in detail.

\subsection{Mesh Processing}
The inherent irregularity and disorder of 3D meshes necessitate a repartitioning of the mesh, requiring the feature vectors within each patch to be arranged in a consistent order~\cite{pang2022masked}. 
To address this, our patch processing consists of two stages: mesh patch splitting and patch embedding. 
Specifically, each triangular face is described by a 13-dimensional feature vector. This vector incorporates the face area, its three normals, its three interior angles, the coordinates of its center point, and the interior product of the multiplication of the face normals and vertex normals.

\textbf{Mesh Patch Splitting:}
Unlike conventional 2D image patch division with fixed size, 3D mesh is composed of irregular face data and disordered faces. It is unreasonable to divide multiple irregular faces into one patch.
We adopt the SubdivNet MAP algorithm \cite{hu2022subdivision}, which applies a remeshing operation to regularize the original mesh through uniform simplification and hierarchical subdivision. 
The visualization of remesh operation is shown in Fig \ref{remesh_ope}.
In detail, each original mesh is first simplified into $F_n$ faces. At each subdivision step, every face is divided into four, yielding $4F_n$ faces after one iteration. After $K$ iterations, the mesh contains $4^K F_n$ faces.
Each patch maintains a consistent ordering of faces, whose 13-dim feature vectors are concatenated to yield a unified representation.

\textbf{Mesh Patch Embedding and Masking:}
After the remeshing operation, the 13-dim feature vectors of all faces within a patch are concatenated to form its patch feature.
To ensure consistency, the faces in each patch are arranged in a fixed order. Each patch representation is then projected through an MLP layer to obtain $H=\{h_i\}^{f_n}_{i=1}$, where $f_n$ denotes the number of patches.
The positional embedding is represented by the center coordinates of each patch, defined as $P=\{p_i\}^{f_n}_{i=1}$. Finally, the input embedding is constructed by combining the patch embedding and positional embedding, denoted as $X=\{x_i\}^{f_n}_{i=1}$. 

\subsection{Reconstruction Branch}
The reconstruction branch comprises a shared mesh encoder and a corresponding mesh decoder. 
Its core function is to extract mesh features from masked oral scan data and uses a decoder to restore the texture, enabling the encoder to learn oral structure information. 
Next, we will introduce these components in detail.

\subsubsection{Encoder and Decoder}
The encoder of the reconstruction branch consists of 12 standard Transformer blocks.
It takes the embedding $X_e\in \mathbb{R}^{(1-r)f_n\times d}$ as input.
$X_e$ is obtained by masking $X$.
The encoder's output is directly fed into the decoder.
The decoder consists of 6 standard Transformer blocks and takes as input both the visible embeddings and mask embedding, along with their positional embeddings. The mask embeddings are shared and learnable.

\subsubsection{Reconstruction Target}
The reconstruction target is defined to recover both the patch vertices and the face features from $X_e$.
Specifically, each masked patch in $X_e$ contains 45 vertices, these vertices are reconstructed by predicting the relative coordinates (the coordinates of the vertices relative to the center point of the patch). 
We adopt the $l_2$-form of Chamfer distance as the reconstruction loss to supervise the predicted vertex relative coordinates $V_p$ and the ground-truth relative coordinates $G_q$. The Chamfer distance loss $L_{CD}$ is formulated as:
\begin{equation}
\begin{split}
L_{CD}(V_p, G_q) = &\frac{1}{|V_p|}\sum_{p \in V_p}\min_{q \in G_q} ||p - q|| \\
&+\frac{1}{|G_q|}\sum_{q \in G_q} \min_{p \in V_p} ||q-p||.
\end{split}
\end{equation}
However, recovering facial information solely from vertex positions is insufficient.
To more effectively reconstruct the shape of the face, we add a linear layer after the decoder to obtain the predicted face features for each patch.
An additional mean squared error (MSE) loss is introduced to supervise the reconstructed face feature $\hat{f}_i$, defined as:
\begin{equation}
L_{MSE}= \frac{1}{|M|} \sum_{i\in M} \| f_i - \hat{f}_i \|_2^2
\end{equation}
where $f_i$ denotes the ground-truth face feature of the i-th masked patch, $\hat{f}_i$ is the corresponding predicted feature, and $M$ is the index set of masked patches.
The final reconstruction loss is the sum of $L_{CD}$ and $L_{MSE}$:
\begin{equation}
L_{re} = L_{CD} + \eta L_{MSE},
\end{equation}
where $\eta$ is the loss weight of face features. 

\subsection{Regression Branch}
The regression branch consists of an encoder, a text-conditioned prompt (TCP) module, and three regression heads.
Its core function is to integrate the information of implant systems and series into the network and predict the abutment parameter. 
Next, we will introduce these components in detail.


\subsubsection{Text-Conditioned Prompt Module}
During abutment design, dentists first determine the implant location according to the patient's information.
Then, they specify the type of implant abutment according to the implant system and series.
This information (implant location, implant system, and series) can effectively guide the abutment design, constraining the range of abutment parameters.
Therefore, integrating this prior information into our prediction network enables the model to perform better in abutment design.
Recent works have shown that text embeddings generated by CLIP can effectively guide the network to focus on the implant area~\cite{yang2023tceip, yang2023tcslot}.
Inspired by these successful applications, we design a text-conditioned prompt (TCP) module to introduce the prior information for the prediction network. 
The architecture of TCP is given in Fig. \ref{Overall_Framework}. 
Specifically, we combine the implant location, implant system, and series into a fixed text template $X_t$, e.g., 'Bottom-45 OSSTEM R'.
Then, $X_t$ is fed into the text encoder of CLIP to generate text embedding $\mathcal{F}_t$:
\begin{equation}
	\mathcal{F}_t = CLIP(X_t), \mathcal{F}_t\in\mathbb{R}^{1\times c},
\end{equation}	

Since the text features extracted by CLIP differ significantly from those learned by the network, we use a fully connected layer to map $\mathcal{F}_t$ to the same feature space as the network features:
\begin{equation}
	\mathcal{F}_{pj} = Linear(unsqueeze(\mathcal{F}_t)), \mathcal{F}_{pj}\in\mathbb{R}^{1\times d}.
\end{equation}

Furthermore, we apply cross-attention to enhance the cross-modal fusion between text and mesh features. 
In detail, mesh features are regarded as the query, while text features serve as the key and value:
\begin{equation}
\mathcal{F}_{ca} = \text{softmax}\left(\frac{\mathcal{F}_{e}\mathcal{F}_{pj}^T}{\sqrt{d}}\right)\mathcal{F}_{pj}, \\
\end{equation}
where $d$ is the feature dimension of $\mathcal{F}_{e}$.
Through the cross-modal fusion, text descriptions can effectively constrain the abutment parameter predictions of the network.

To further highlight the characteristics of $\mathcal{F}_{ca}$, we employ max pooling and mean pooling to produce $\mathcal{F}_{max}$ and $\mathcal{F}_{mean}$. These two pooled features are then concatenated and fed into a fully connected layer to generate $\mathcal{F}_{o}$:
\begin{equation}
	\mathcal{F}_{o} = \text{FC}(concat(\mathcal{F}_{max}, \mathcal{F}_{mean})).
\end{equation}
$\mathcal{F}_{o}$ is the final output of the TCP module, which is fed to the regression heads for abutment parameter prediction.

\subsubsection{Encoder and Regression Heads}
The encoder of the regression branch is consistent with that of the reconstruction branch.
By sharing knowledge of oral structure learned from the reconstruction branch, the encoder of the regression branch continues to learn fine-grained features about abutment design through supervised learning. 
Unlike the reconstruction branch, the regression branch encoder takes complete patch embeddings $X$ as input, which does not involve masking. 
The encoded mesh features $\mathcal{F}_e$ are represented as:
\begin{equation}
	\mathcal{F}_e = Encoder(X), \mathcal{F}_e\in\mathbb{R}^{f_n\times d},
\end{equation}
where $f_n$ represents the number of patches, $d$ represents the mesh embedding dimension.

The decoder consists of three fully connected layers followed by three independent regression heads. The fully connected layer reduces the fused feature dimension and feeds the resulting features into the regression heads for abutment parameter prediction.
We use the mean squared error (MSE) loss $ L_{MSE}^{\star}$ and the smooth L1 loss $L_{l1}$ 
to supervise the regression head:
\begin{equation}
	L_{l1}(x_i, y_i)=
	\begin{cases} 
	0.5 \cdot (x_i-y_i)^2 / \hbar, & |x_i-y_i| < \hbar \\
	|x_i-y_i| - 0.5 \cdot \hbar, &\text{otherwise},
	\end{cases}
\end{equation}
\begin{equation}
	 L_{MSE}^{\star}(x_i, y_i)=\frac{1}{N} \sum_{i=1}^{N} (x_i - y_i )^2
\end{equation}
\begin{equation}
	L_{rg} = L_{l1} + L_{MSE}^{\star},
\end{equation}
where $x_i$ is the i-th predicted value and $y_i$ is the i-th ground-truth value, $N$ represents the total number of samples, and $\hbar$ is the smoothing function. 
Finally, the total training loss is the combination of $L_{re}$ and $L_{rg}$:
\begin{equation}
	L_{total} = \varsigma L_{re} + L_{rg},
\end{equation}
where $\varsigma$ is a weighting factor for the reconstruction loss.

\begin{figure*}\centering
\includegraphics[scale=0.25]{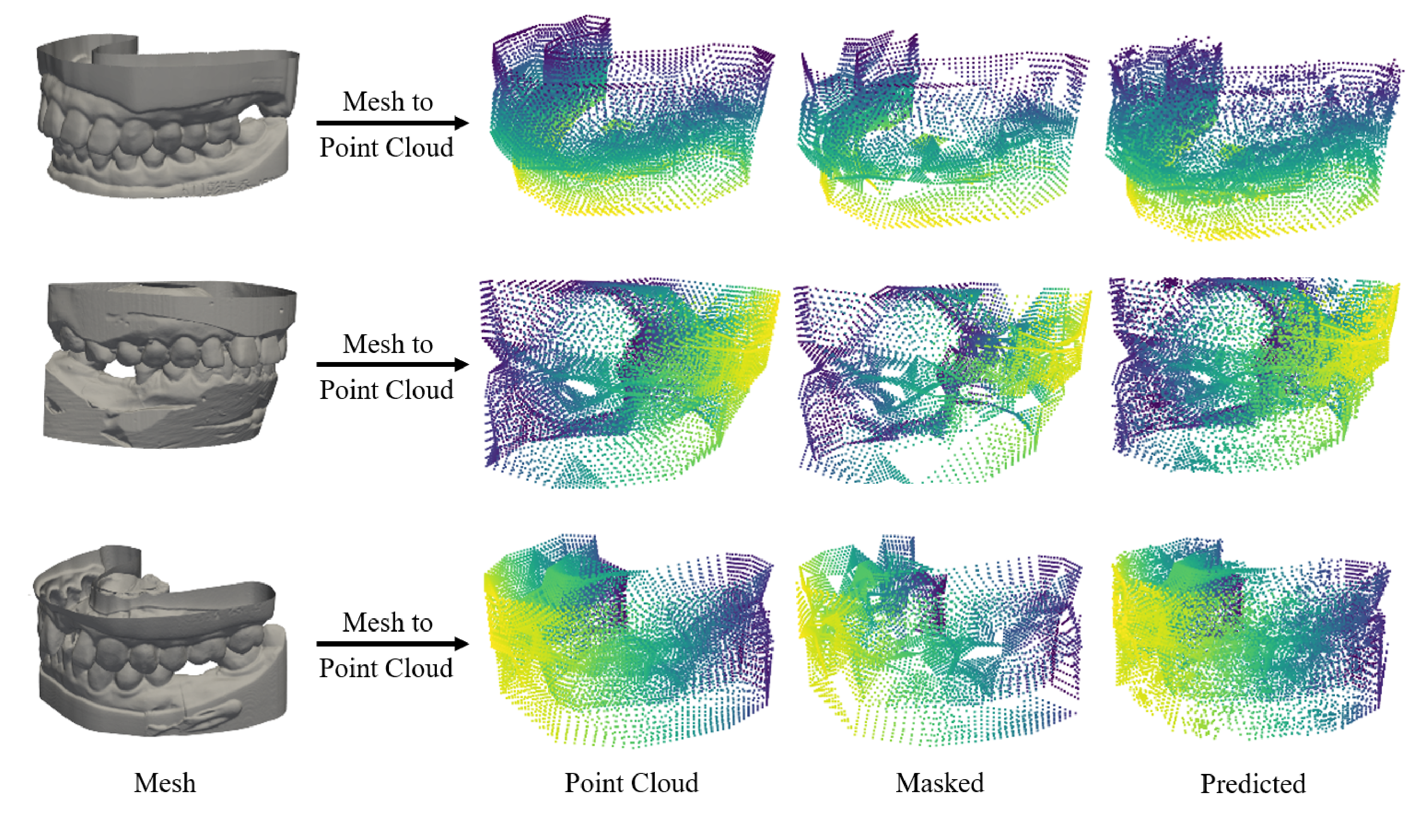}
\caption{Visualization of the reconstruction branch.}
\label{rec_vis}
\end{figure*}

\begin{figure}\centering
\includegraphics[scale=0.26]{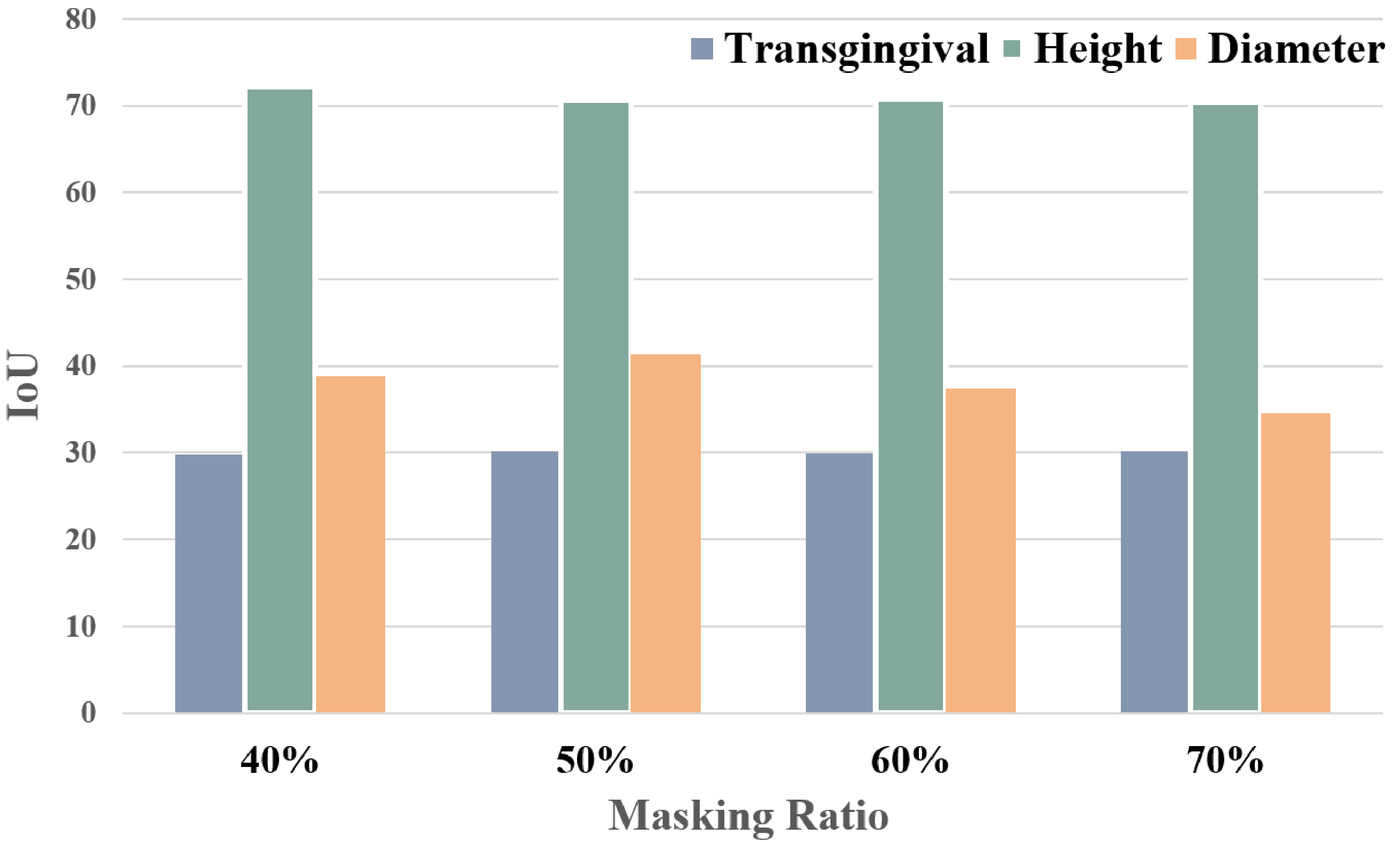}
\caption{The IoU performance of the reconstruction branch with different masking ratios.}
\label{masking_ratio}
\end{figure}

\section{Experiments}
\label{experiments}
\subsection{DataSets}
To validate the effectiveness of the proposed SS$A^3$D model, a dataset comprising 8,737 intraoral scans, each featuring a single missing tooth, was collected. 
All annotations were performed by nine senior dental technicians. For each case, the technicians determined the appropriate abutment type by manually measuring the intraoral scan. The three key parameters corresponding to the selected abutment type served as the ground-truth labels. To ensure a balanced representation of all abutment categories, 85\% of the samples from each type were allocated to the training set, with the remaining 15\% reserved for testing.
To ensure a standardized data format, all intraoral scan data underwent a comprehensive preprocessing pipeline. 
This pipeline consisted of three key steps: (1) a manifold operation was first applied to ensure the geometric integrity of each mesh; (2) the resulting models were then simplified to a uniform density of 500 faces; and (3) finally, the MAPS algorithm was employed to remesh the data. 
Through this process, each mesh was subdivided three times, ultimately generating high-quality meshes with 32,000 faces suitable for subsequent analysis.

\subsection{Evaluation Criteria}
The standard clinical protocol involves measuring the implant site to determine key abutment parameters, based on which a stock abutment with the closest matching specifications is selected. 
The parameter set of this clinically chosen abutment thus constitutes the ground truth against which the predictions generated by the SS$A^3$D are evaluated. 
For the quantitative assessment of prediction accuracy, the deviation between the predicted and true values is measured using the Intersection over Union (IoU) metric. 
This is implemented by representing each parameter as a fixed unit area and computing the IoU between the predicted and the ground truth areas. 
The metric is defined as follows:
\begin{equation}
	\text{IoU}(pv_j, gt_j) = \frac{|pv_j\cap gt_j|}{|pv_j\cup gt_j|}
\end{equation}
where $pv_j\in[x, x+1]$ denotes the predicted area of the $j$-th predicted value, and $gt_j\in[y, y+1]$ denotes the corresponding ground-truth area. 
When the IoU reaches 0.42, the difference between the predicted and actual values is approximately 0.4 mm, which is within the acceptable tolerance range for clinical applications.

\begin{table*}[]  
\centering
\caption{Ablation experiments on the TCP module and reconstruction branch.}
\begin{tabular}{c|c|ccc}
\hline
	 Reconstruction Branch & TCP & {Transgingival} & {Diameter} & {Height} \\
	   \hline 
	  \cxmark & \ding{52}  & 29.24 & 69.99 & 31.94 \\
	 \ding{52} & \cxmark  & 29.85 & 62.94 & 20.86 \\
	\ding{52} & \ding{52} & \textbf{30.58} & \textbf{70.69} & \textbf{41.41} \\ \hline
	\end{tabular}
	\label{tcp_module}
\end{table*}

\subsection{Ablation Studies}
\subsubsection{Ablation Study of TCP Module and Reconstruction Branch}
To validate the effectiveness of the proposed TCP module, we conducted ablation studies comparing SS$A^3$D with and without it. As shown in Table~\ref{tcp_module}, incorporating the TCP module consistently improves the IoU for the transgingival, diameter, and height parameters, with the most substantial gain observed for height. This confirms that the TCP module successfully integrates textual guidance with mesh features, thereby enhancing the model's regression capability.

The contribution of the reconstruction branch is also significant. Its removal results in performance degradation across all three parameters, with the height IoU being the most adversely affected. This finding indicates that the auxiliary reconstruction task facilitates more robust mesh feature learning, which in turn improves the primary prediction performance of SS$A^3$D.

\begin{table*}[]
\centering
\caption{Comparison of Query Settings in TCP Module.}
\begin{tabular}{ccc|ccc|c}
\hline
\multicolumn{3}{c|}{\textbf{TCP}} & \multirow{2}{*}{Transgingival} & \multirow{2}{*}{Diameter} & \multirow{2}{*}{Height} & \multirow{2}{*}{Mean}  \\ \cline{1-3}
Q         & K         & V        &                        &                      &                      \\ \hline
Mesh      & Text      & Text     & 30.58 & \textbf{70.69} & \textbf{41.41} & \textbf{47.44} \\
Text      & Mesh      & Mesh     & \textbf{30.68} & 70.36 & 38.18 & 46.41 \\ \hline
\end{tabular}
	\label{tcp_qkv}
\end{table*}

\subsubsection{Ablation Study of Masking Ratio in Reconstruction Branch}
The masking ratio in the reconstruction branch is a key factor governing the model's capacity to learn mesh features. To determine the optimal setting, we conducted a comparative analysis with varying masking ratios while maintaining a fixed reconstruction loss ratio of 0.1. As illustrated in Figure~\ref{masking_ratio}, a 40\% ratio yields the highest IoU for height, while a 50\% ratio achieves the best performance for the transgingival and diameter parameters. 
Based on these findings, a ratio of 50\% was selected as it provides the most balanced performance across all parameters, thereby enabling the model to effectively accomplish the automatic abutment design task.

\subsubsection{Comparison of Different Query Settings in TCP}
The choice of query features in the TCP module dictates the mode of cross-modal fusion. To investigate this, we conducted experiments using different modal features as queries (Table~\ref{tcp_qkv}). The results indicate that employing mesh features as the query yields superior performance for the diameter and height parameters, although not for transgingival penetration. To achieve more well-balanced results across all parameters, we ultimately selected mesh features as the query for the automatic abutment design task.

\begin{table}[]
\centering
\caption{Effect of training data size on the IoU performance of the SS$A^3$D model}
\begin{tabular}{c|ccccc}
\hline
	\multirow{2}{*}{Parameter} & \multicolumn{5}{c}{Training data size} \\
	  \cline{2-6}
	 & 20\% & 40\% & 60\% & 80\% & 100\% \\ \hline
	Transgingival & 29.75 & 28.80 & 30.22 & 29.94 & \textbf{30.58} \\ 
	Diameter & 69.92 & 71.32 & 71.43 & \textbf{71.55} & 70.69 \\
	Height & 36.17 & 33.51 & 41.39 & 41.32 & \textbf{41.41}\\
	 \hline
	\end{tabular}
	\label{Datasize_compare}
\end{table}

\begin{table*}[]  
\centering
\caption{Comparison of IoU performance and training time of different paradigms of SS$A^3$D model}
\begin{tabular}{c|ccc|c}
\hline
	 Paradigm & Transgingival & Diameter & Height & Training Time (hours)\\ \hline
	SSL+FT & 29.75 & 70.05 & 32.77 & 5.9 + 1.8 \\ 
	SSAT & \textbf{30.58} & \textbf{70.69} & \textbf{41.41} & \textbf{3.1} ($\downarrow$ 4.6) \\
	 \hline
	\end{tabular}
	\label{time_compare}
\end{table*}

\subsubsection{Comparison of Different Training Data Size}
To demonstrate that SS$A^3$D performs well even with a limited number of training data, we constructed training sets with varying proportions to assess its performance.
Table \ref{Datasize_compare} presents the IoU performance of the proposed SS$A^3$D model trained with different proportions of the training data. The SS$A^3$D model configurations remain identical across experiments, except for the number of training samples, and the IoU values are evaluated on the same test set.
As shown in Table \ref{Datasize_compare}, even when only 20\% of the original training data is used, the model still achieves satisfactory IoU performance for the transgingival and diameter parameters, although the IoU for the height parameter decreases. 
When the proportion of the training set increased to 40\%, the SS$A^3$D model exhibited the most significant decline in height parameter performance, with the IoU decreasing by 7.9\%, yet remaining within an acceptable error range.
The best overall performance is observed when the full training dataset is employed.
In summary, these results indicate that the proposed SS$A^3$D model exhibits strong feature representation and generalization capability, even under limited data conditions.

\subsubsection{Visualization of Reconstruction Branch}
To evaluate the feature-learning capability of the SS$A^3$D reconstruction branch, we tasked it with restoring oral scan meshes where 50\% of the faces were masked. 
As shown in Fig.~\ref{rec_vis}, the branch faithfully reconstructs the missing portions to align with the original geometry. Although some fine local details are not perfectly recovered, the overall structural integrity is maintained. These results confirm that the reconstruction branch effectively learns meaningful representations of the oral scan data.

\begin{table*}[]
\centering
\caption{Comparison of SS$A^3$D with other mainstream.}
\begin{tabular}{c|c|c|ccc}
\hline
	Input & Pre-training & Method & Transgingival & Diameter & Height \\ \hline 
	& \cxmark & PointNet & 28.61 & 46.18 & 22.81 \\
     & \cxmark & PointNet++  & 29.14 & 63.26 & 24.29 \\
	Point Cloud & \cxmark & PointFormer & 29.37 & 63.89 & 19.91 \\
     & \ding{52} & PointMAE  & 29.14 & 58.57 & 17.76 \\
     & \ding{52} & PointMamba  & 28.85 & 59.72 & 14.67 \\
     & \ding{52} & PointFEMAE & 30.15 & 62.82 & 15.60 \\ \hline
     \multirow{2}{*}{Mesh} & \ding{52} & MeshMAE & 29.55 & 62.86 & 17.29 \\
	& \cxmark & SS$A^{3}$D & \textbf{30.58} & \textbf{70.69} & \textbf{41.41} \\ \hline
	\end{tabular}
  \label{Diff_model}
\end{table*}

\subsubsection{Comparison with the SSL methods}
To compare the effectiveness of the traditional SSL and SSAT paradigms, we evaluated both prediction accuracy and training efficiency. The experimental setup is as follows: the SSL+FT paradigm uses the reconstruction branch of the proposed SS$A^3$D model for pre-training, after which the regression branch is separately trained and fine-tuned. In contrast, the SSAT paradigm directly trains the complete SS$A^3$D model, with the reconstruction and regression branches being jointly optimized; during inference, only the regression branch is utilized.

All models were trained for an equal number of epochs, including a 300-epoch pre-training phase for SSL, to ensure a fair comparison. The results, summarized in Table~\ref{time_compare}, demonstrate that the proposed SS$A^3$D under the SSAT paradigm not only achieves superior accuracy—notably in height prediction—but also reduces the training time by more than half compared to the traditional SSL approach. This two-fold improvement in efficiency, simultaneously with enhanced performance, underscores the significant advantages of our method.

\subsubsection{Comparison to State-of-the-art Methods}
To further validate the superiority of the proposed SS$A^3$D, we conducted a comparative analysis against several state-of-the-art methods. The benchmarked approaches include point cloud-based methods—PointNet~\cite{qi2017pointnet}, PointNet++~\cite{qi2017pointnet++}, PointFormer~\cite{chen2022pointformer}, PointMAE~\cite{pang2022masked}, PointMamba~\cite{liang2024pointmamba}, and PointFEMAE~\cite{zha2024towards}—as well as the mesh-based method MeshMAE~\cite{liang2022meshmae}. 
The IoU results for the three evaluated parameters are summarized in Table \ref{Diff_model}.
The proposed SS$A^3$D achieves the best overall performance across all parameters. A particularly noteworthy finding is that our method yields superior results, especially for the height parameter (its accuracy is nearly 2-3 times that of other methods), even when the pre-training and training datasets are identical—a scenario where pre-trained models typically do not outperform their non-pre-trained counterparts. This result underscores the exceptional capability of SS$A^3$D in effectively capturing discriminative 3D mesh features

\section{Conclusion}
In this paper, we propose a self-supervised assisted automatic abutment design framework (SS$A^3$D), which simplifies the abutment design process and performs well with limited labeled data. 
It employs a dual-branch architecture with a reconstruction branch and a regression branch. 
The reconstruction branch learns to restore masked intraoral scan data and transfers the learned structural information to the regression branch. 
The regression branch then predicts the abutment parameters under supervision. 
We design a Text-Conditioned Prompt (TCP) module to incorporate clinical information (such as implant location, system, and series) into SS$A^3$D, guiding the network to focus on relevant regions and constraining the parameter predictions. 
Extensive experiments on a collected dataset show that SS$A^3$D achieves state-of-the-art performance compared to other methods, significantly improving the accuracy and efficiency of automated abutment design.


\bibliographystyle{model5-names}
\bibliography{refs}

@article{hu2022subdivision,
  title={Subdivision-based mesh convolution networks},
  author={Hu, Shi-Min and Liu, Zheng-Ning and Guo, Meng-Hao and Cai, Jun-Xiong and Huang, Jiahui and Mu, Tai-Jiang and Martin, Ralph R},
  journal={ACM Transactions on Graphics (TOG)},
  volume={41},
  number={3},
  pages={1--16},
  year={2022},
  publisher={ACM New York, NY}
}

@inproceedings{yang2024simplify,
  title={Simplify implant depth prediction as video grounding: A texture perceive implant depth prediction network},
  author={Yang, Xinquan and Li, Xuguang and Luo, Xiaoling and Zeng, Leilei and Zhang, Yudi and Shen, Linlin and Deng, Yongqiang},
  booktitle={International Conference on Medical Image Computing and Computer-Assisted Intervention},
  pages={606--615},
  year={2024},
  organization={Springer}
}

@article{yang2024implantformer,
  title={ImplantFormer: vision transformer-based implant position regression using dental CBCT data},
  author={Yang, Xinquan and Li, Xuguang and Li, Xuechen and Wu, Peixi and Shen, Linlin and Deng, Yongqiang},
  journal={Neural Computing and Applications},
  volume={36},
  number={12},
  pages={6643--6658},
  year={2024},
  publisher={Springer}
}

@article{yang2024two,
  title={Two-stream regression network for dental implant position prediction},
  author={Yang, Xinquan and Li, Xuguang and Li, Xuechen and Chen, Wenting and Shen, Linlin and Li, Xin and Deng, Yongqiang},
  journal={Expert Systems with Applications},
  volume={235},
  pages={121135},
  year={2024},
  publisher={Elsevier}
}

@inproceedings{yang2023tceip,
  title={Tceip: Text condition embedded regression network for dental implant position prediction},
  author={Yang, Xinquan and Xie, Jinheng and Li, Xuguang and Li, Xuechen and Li, Xin and Shen, Linlin and Deng, Yongqiang},
  booktitle={International Conference on Medical Image Computing and Computer-Assisted Intervention},
  pages={317--326},
  year={2023},
  organization={Springer}
}

@inproceedings{yang2023tcslot,
  title={Tcslot: Text guided 3d context and slope aware triple network for dental implant position prediction},
  author={Yang, Xinquan and Xie, Jinheng and Li, Xuechen and Li, Xuguang and Shen, Linlin and Deng, Yongqiang},
  booktitle={2023 IEEE International Conference on Bioinformatics and Biomedicine (BIBM)},
  pages={726--732},
  year={2023},
  organization={IEEE}
}

@article{krenmayr2025evaluating,
  title={Evaluating masked self-supervised learning frameworks for 3D dental model segmentation tasks},
  author={Krenmayr, Lucas and von Schwerin, Reinhold and Schaudt, Daniel and Riedel, Pascal and Hafner, Alexander and Geserick, Marc},
  journal={Scientific Reports},
  volume={15},
  number={1},
  pages={16818},
  year={2025},
  publisher={Nature Publishing Group UK London}
}

@inproceedings{almalki2024self,
  title={Self-supervised learning with masked autoencoders for teeth segmentation from intra-oral 3d scans},
  author={Almalki, Amani and Latecki, Longin Jan},
  booktitle={Proceedings of the IEEE/CVF Winter Conference on Applications of Computer Vision},
  pages={7820--7830},
  year={2024}
}

@inproceedings{qi2017pointnet,
  title={Pointnet: Deep learning on point sets for 3d classification and segmentation},
  author={Qi, Charles R and Su, Hao and Mo, Kaichun and Guibas, Leonidas J},
  booktitle={Proceedings of the IEEE conference on computer vision and pattern recognition},
  pages={652--660},
  year={2017}
}

@article{qi2017pointnet++,
  title={Pointnet++: Deep hierarchical feature learning on point sets in a metric space},
  author={Qi, Charles Ruizhongtai and Yi, Li and Su, Hao and Guibas, Leonidas J},
  journal={Advances in neural information processing systems},
  volume={30},
  year={2017}
}

@inproceedings{pang2022masked,
  title={Masked autoencoders for point cloud self-supervised learning},
  author={Pang, Yatian and Wang, Wenxiao and Tay, Francis EH and Liu, Wei and Tian, Yonghong and Yuan, Li},
  booktitle={European conference on computer vision},
  pages={604--621},
  year={2022},
  organization={Springer}
}

@inproceedings{chen2022pointformer,
  title={Pointformer: A dual perception attention-based network for point cloud classification},
  author={Chen, Yijun and Yang, Zhulun and Zheng, Xianwei and Chang, Yadong and Li, Xutao},
  booktitle={Proceedings of the Asian Conference on Computer Vision},
  pages={3291--3307},
  year={2022}
}

@inproceedings{liang2022meshmae,
  title={Meshmae: Masked autoencoders for 3d mesh data analysis},
  author={Liang, Yaqian and Zhao, Shanshan and Yu, Baosheng and Zhang, Jing and He, Fazhi},
  booktitle={European conference on computer vision},
  pages={37--54},
  year={2022},
  organization={Springer}
}

@inproceedings{zha2024towards,
  title={Towards compact 3d representations via point feature enhancement masked autoencoders},
  author={Zha, Yaohua and Ji, Huizhen and Li, Jinmin and Li, Rongsheng and Dai, Tao and Chen, Bin and Wang, Zhi and Xia, Shu-Tao},
  booktitle={Proceedings of the AAAI Conference on Artificial Intelligence},
  volume={38},
  number={7},
  pages={6962--6970},
  year={2024}
}

@article{tartea2023comparative,
  title={Comparative study of dental custom CAD-CAM implant abutments and dental implant stock abutments},
  author={T{\^a}rtea, Daniel Adrian and Ionescu, Mihaela and Manolea, Horia Octavian and Mercuț, Veronica and Ob{\u{a}}dan, Eugenia and Am{\u{a}}r{\u{a}}scu, Marina Olimpia and M{\u{a}}r{\u{a}}șescu, Petre Costin and D{\u{a}}guci, Luminița and Popescu, Sanda Mihaela},
  journal={Journal of Clinical Medicine},
  volume={12},
  number={6},
  pages={2128},
  year={2023},
  publisher={MDPI}
}

@article{benakatti2021dental,
  title={Dental implant abutments and their selection—A review},
  author={Benakatti, Veena and Sajjanar, Jayashree Arun and Acharya, Aditya Ramnarayan},
  journal={J. Evol. Med. Dent. Sci},
  volume={10},
  pages={3053--3059},
  year={2021}
}

@article{cui2021tsegnet,
  title={TSegNet: An efficient and accurate tooth segmentation network on 3D dental model},
  author={Cui, Zhiming and Li, Changjian and Chen, Nenglun and Wei, Guodong and Chen, Runnan and Zhou, Yuanfeng and Shen, Dinggang and Wang, Wenping},
  journal={Medical Image Analysis},
  volume={69},
  pages={101949},
  year={2021},
  publisher={Elsevier}
}

@article{jang2021fully,
  title={A fully automated method for 3D individual tooth identification and segmentation in dental CBCT},
  author={Jang, Tae Jun and Kim, Kang Cheol and Cho, Hyun Cheol and Seo, Jin Keun},
  journal={IEEE transactions on pattern analysis and machine intelligence},
  volume={44},
  number={10},
  pages={6562--6568},
  year={2021},
  publisher={IEEE}
}

@inproceedings{qiu2022darch,
  title={Darch: Dental arch prior-assisted 3d tooth instance segmentation with weak annotations},
  author={Qiu, Liangdong and Ye, Chongjie and Chen, Pei and Liu, Yunbi and Han, Xiaoguang and Cui, Shuguang},
  booktitle={Proceedings of the IEEE/CVF Conference on Computer Vision and Pattern Recognition},
  pages={20752--20761},
  year={2022}
}

@article{xu20183d,
  title={3D tooth segmentation and labeling using deep convolutional neural networks},
  author={Xu, Xiaojie and Liu, Chang and Zheng, Youyi},
  journal={IEEE transactions on visualization and computer graphics},
  volume={25},
  number={7},
  pages={2336--2348},
  year={2018},
  publisher={IEEE}
}

@article{tian2021efficient,
  title={Efficient computer-aided design of dental inlay restoration: a deep adversarial framework},
  author={Tian, Sukun and Wang, Miaohui and Yuan, Fulai and Dai, Ning and Sun, Yuchun and Xie, Wuyuan and Qin, Jing},
  journal={IEEE Transactions on Medical Imaging},
  volume={40},
  number={9},
  pages={2415--2427},
  year={2021},
  publisher={IEEE}
}

@article{shen2023transdfnet,
  title={TranSDFNet: Transformer-based truncated signed distance fields for the shape design of removable partial denture clasps},
  author={Shen, Xinze and Zhang, Changdong and Jia, Xiuyi and Li, Dawei and Liu, Tingting and Tian, Sukun and Wei, Wei and Sun, Yuchun and Liao, Wenhe},
  journal={IEEE Journal of Biomedical and Health Informatics},
  volume={27},
  number={10},
  pages={4950--4960},
  year={2023},
  publisher={IEEE}
}

@article{yuan2020personalized,
  title={Personalized design technique for the dental occlusal surface based on conditional generative adversarial networks},
  author={Yuan, Fulai and Dai, Ning and Tian, Sukun and Zhang, Bei and Sun, Yuchun and Yu, Qing and Liu, Hao},
  journal={International Journal for Numerical Methods in Biomedical Engineering},
  volume={36},
  number={5},
  pages={e3321},
  year={2020},
  publisher={Wiley Online Library}
}

@article{krishnan2022self,
  title={Self-supervised learning in medicine and healthcare},
  author={Krishnan, Rayan and Rajpurkar, Pranav and Topol, Eric J},
  journal={Nature Biomedical Engineering},
  volume={6},
  number={12},
  pages={1346--1352},
  year={2022},
  publisher={Nature Publishing Group UK London}
}

@article{zhang2017method,
  title={A method for using solid modeling CAD software to create an implant library for the fabrication of a custom abutment},
  author={Zhang, Jing and Zhang, Rimei and Ren, Guanghui and Zhang, Xiaojie},
  journal={The Journal of prosthetic dentistry},
  volume={117},
  number={2},
  pages={209--213},
  year={2017},
  publisher={Elsevier}
}

@article{liu2022hierarchical,
  title={Hierarchical self-supervised learning for 3D tooth segmentation in intra-oral mesh scans},
  author={Liu, Zuozhu and He, Xiaoxuan and Wang, Hualiang and Xiong, Huimin and Zhang, Yan and Wang, Gaoang and Hao, Jin and Feng, Yang and Zhu, Fudong and Hu, Haoji},
  journal={IEEE Transactions on Medical Imaging},
  volume={42},
  number={2},
  pages={467--480},
  year={2022},
  publisher={IEEE}
}

@inproceedings{das2024limited,
  title={Limited data, unlimited potential: A study on vits augmented by masked autoencoders},
  author={Das, Srijan and Jain, Tanmay and Reilly, Dominick and Balaji, Pranav and Karmakar, Soumyajit and Marjit, Shyam and Li, Xiang and Das, Abhijit and Ryoo, Michael S},
  booktitle={Proceedings of the IEEE/CVF winter conference on applications of computer vision},
  pages={6878--6888},
  year={2024}
}

@article{liu2021efficient,
  title={Efficient training of visual transformers with small datasets},
  author={Liu, Yahui and Sangineto, Enver and Bi, Wei and Sebe, Nicu and Lepri, Bruno and Nadai, Marco},
  journal={Advances in Neural Information Processing Systems},
  volume={34},
  pages={23818--23830},
  year={2021}
}

@article{cai2024ssad,
  title={SSAD: Self-supervised Auxiliary Detection Framework for Panoramic X-ray based Dental Disease Diagnosis},
  author={Cai, Zijian and Yang, Xinquan and Li, Xuguang and Luo, Xiaoling and Li, Xuechen and Shen, Linlin and Meng, He and Deng, Yongqiang},
  journal={arXiv preprint arXiv:2406.13963},
  year={2024}
}

@article{abichandani2013abutment,
  title={Abutment selection, designing, and its influence on the emergence profile: A comprehensive review},
  author={Abichandani, Sagar J and Nadiger, Ramesh and Kavlekar, Abhishek S and others},
  journal={European Journal of Prosthodontics},
  volume={1},
  number={1},
  pages={1},
  year={2013},
  publisher={Medknow Publications}
}

@article{shah2023literature,
  title={A literature review on implant abutment types, materials, and fabrication processes},
  author={Shah, Khushali K and Sivaswamy, Vinay},
  journal={Journal of long-term effects of medical implants},
  volume={33},
  number={1},
  year={2023},
  publisher={Begel House Inc.}
}

@article{priest2005virtual,
  title={Virtual-designed and computer-milled implant abutments},
  author={Priest, George},
  journal={Journal of Oral and Maxillofacial Surgery},
  volume={63},
  number={9},
  pages={22--32},
  year={2005},
  publisher={Elsevier}
}

@article{kang2020abutment,
  title={Abutment-Level Digital Impression Using Abutment Matching Algorithm, And Insurance-Covered Implant Prosthesis By Metal 3D Printing: A Case Report},
  author={Kang, In-Ho and Park, Ji-Man and Shim, June-Sung},
  journal={Journal of implantology and applied sciences},
  volume={24},
  number={3},
  pages={137--147},
  year={2020},
  publisher={The Korean Academy Of Oral \& Maxillofacial Implantology}
}

@inproceedings{chen2020simple,
  title={A simple framework for contrastive learning of visual representations},
  author={Chen, Ting and Kornblith, Simon and Norouzi, Mohammad and Hinton, Geoffrey},
  booktitle={International conference on machine learning},
  pages={1597--1607},
  year={2020},
  organization={PmLR}
}

@inproceedings{he2020momentum,
  title={Momentum contrast for unsupervised visual representation learning},
  author={He, Kaiming and Fan, Haoqi and Wu, Yuxin and Xie, Saining and Girshick, Ross},
  booktitle={Proceedings of the IEEE/CVF conference on computer vision and pattern recognition},
  pages={9729--9738},
  year={2020}
}

@article{bao2021beit,
  title={Beit: Bert pre-training of image transformers},
  author={Bao, Hangbo and Dong, Li and Piao, Songhao and Wei, Furu},
  journal={arXiv preprint arXiv:2106.08254},
  year={2021}
}

@article{ma2025multi,
  title={Multi-Encoding Contrastive Learning for Dual-Stream Self-Supervised 3D Dental Segmentation Network},
  author={Ma, Tian and Wei, Xiaoyuan and Zhai, Jiechen and Zhang, Ziang and Li, Yawen and Li, Yuancheng},
  journal={Technologies},
  volume={13},
  number={9},
  pages={419},
  year={2025},
  publisher={MDPI}
}

@inproceedings{yu2022point,
  title={Point-bert: Pre-training 3d point cloud transformers with masked point modeling},
  author={Yu, Xumin and Tang, Lulu and Rao, Yongming and Huang, Tiejun and Zhou, Jie and Lu, Jiwen},
  booktitle={Proceedings of the IEEE/CVF conference on computer vision and pattern recognition},
  pages={19313--19322},
  year={2022}
}

@article{zhang2022point,
  title={Point-m2ae: multi-scale masked autoencoders for hierarchical point cloud pre-training},
  author={Zhang, Renrui and Guo, Ziyu and Gao, Peng and Fang, Rongyao and Zhao, Bin and Wang, Dong and Qiao, Yu and Li, Hongsheng},
  journal={Advances in neural information processing systems},
  volume={35},
  pages={27061--27074},
  year={2022}
}

@article{chen2023pointgpt,
  title={Pointgpt: Auto-regressively generative pre-training from point clouds},
  author={Chen, Guangyan and Wang, Meiling and Yang, Yi and Yu, Kai and Yuan, Li and Yue, Yufeng},
  journal={Advances in Neural Information Processing Systems},
  volume={36},
  pages={29667--29679},
  year={2023}
}

@article{liang2024pointmamba,
  title={Pointmamba: A simple state space model for point cloud analysis},
  author={Liang, Dingkang and Zhou, Xin and Xu, Wei and Zhu, Xingkui and Zou, Zhikang and Ye, Xiaoqing and Tan, Xiao and Bai, Xiang},
  journal={Advances in neural information processing systems},
  volume={37},
  pages={32653--32677},
  year={2024}
}

@article{choi2023influence,
  title={Influence of implant--abutment connection biomechanics on biological response: a literature review on interfaces between implants and abutments of titanium and zirconia},
  author={Choi, Sunyoung and Kang, Young Suk and Yeo, In-Sung Luke},
  journal={Prosthesis},
  volume={5},
  number={2},
  pages={527--538},
  year={2023},
  publisher={MDPI}
}

@article{prpic2025influence,
  title={Influence of Design Parameters on Implant Abutment Performance: A Scoping Review},
  author={Prpic, Vladimir and Kosec, Petar and Skec, Stanko and Catic, Amir},
  journal={Journal of functional biomaterials},
  volume={16},
  number={9},
  pages={342},
  year={2025},
  publisher={MDPI}
}

@article{ramadan2025fully,
  title={Fully digital workflow for a CAD-CAM custom healing abutment with an optimal emergence profile: A dental technique},
  author={Ramadan, Rania E and Razek, Mahmoud Khamis Abdel and Mohamed, Faten S and Abd-Ellah, Mervat E},
  journal={The Journal of Prosthetic Dentistry},
  year={2025},
  publisher={Elsevier}
}

@inproceedings{chen2021exploring,
  title={Exploring simple siamese representation learning},
  author={Chen, Xinlei and He, Kaiming},
  booktitle={Proceedings of the IEEE/CVF conference on computer vision and pattern recognition},
  pages={15750--15758},
  year={2021}
}

@article{zhang2025cmae,
  title={Cmae-3d: Contrastive masked autoencoders for self-supervised 3d object detection},
  author={Zhang, Yanan and Chen, Jiaxin and Huang, Di},
  journal={International Journal of Computer Vision},
  volume={133},
  number={5},
  pages={2783--2804},
  year={2025},
  publisher={Springer}
}

@inproceedings{wei2024towards,
  title={Towards latent masked image modeling for self-supervised visual representation learning},
  author={Wei, Yibing and Gupta, Abhinav and Morgado, Pedro},
  booktitle={European Conference on Computer Vision},
  pages={1--17},
  year={2024},
  organization={Springer}
}

@article{xu2024self,
  title={Self-supervised medical image segmentation using deep reinforced adaptive masking},
  author={Xu, Zhenghua and Liu, Yunxin and Xu, Gang and Lukasiewicz, Thomas},
  journal={IEEE Transactions on Medical Imaging},
  year={2024},
  publisher={IEEE}
}

\end{document}